\documentclass[10pt, a4paper]{article}

\usepackage[final]{lrec2026} 

\usepackage{times}
\usepackage{latexsym}

\usepackage[T1]{fontenc}
\usepackage{cleveref}

\usepackage[utf8]{inputenc}

\usepackage{microtype}

\usepackage{inconsolata}

\usepackage{graphicx}

\usepackage{tikz}

\usetikzlibrary{positioning, shapes.multipart}
\newcommand{\voc}[1]{\texttt{#1}}

\title{SETUP: Sentence-level English-To-Uniform Meaning Representation Parser}

\name{Emma Markle, Javier Gutierrez Bach, Shira Wein} 

\address{Amherst College \\
         \{emarkle26, jgutierrezbach26, swein\}@amherst.edu\\}

\abstract{
Uniform Meaning Representation (UMR) is a novel graph-based semantic representation which captures the core meaning of a text, with flexibility incorporated into the annotation schema such that the breadth of the world's languages can be annotated (including low-resource languages).
While UMR shows promise in enabling language documentation, improving low-resource language technologies, and adding interpretability, the downstream applications of UMR can only be fully explored when text-to-UMR parsers enable the automatic large-scale production of accurate UMR graphs at test time.
Prior work on text-to-UMR parsing is limited to date.
In this paper, we introduce two methods for English text-to-UMR parsing, one of which fine-tunes existing parsers for
Abstract Meaning Representation
and the other, which leverages a converter
from Universal Dependencies, using prior
work as a baseline. Our best-performing model, which we call SETUP, achieves an AnCast score of 84 and a SMATCH++ score of 91, indicating substantial gains towards automatic UMR parsing.
 \\ \newline  }

\begin{document}

\maketitleabstract

\section{Introduction}
\label{sec:intro}

Uniform Meaning Representation (UMR; \citealp{van2021designing}) is a graph-based semantic framework designed to capture the meaning of text across various languages, accounting for linguistic diversity during the annotation process. An example UMR graph can be seen in \Cref{fig:umr_example}. UMR builds on the foundations of Abstract Meaning Representation (AMR; \citealp{banarescu-etal-2013-abstract}), extending its scope to handle multilingual and document-level annotation. Unlike AMR, which was originally developed for English and later adapted cross-lingually \citep{wein-schneider-2024-assessing, shirawein_diss}, UMR is designed as a multilingual semantic framework with a unified annotation schema. Its lattice-based annotation design and Stage 0 procedure further make it well-suited for low- and no-resource languages, offering stronger cross-lingual consistency and broader applicability than AMR.
UMR additionally encodes aspect, modality, scope, and document-level relations.


\begin{figure}[t]
    \small
\centering
\resizebox{\linewidth}{!}{
\begin{tikzpicture}[
blue/.style={rectangle, draw=black, very thick, minimum size=6mm},
]
	\node[blue](s) at (10,10) {\voc{answer-01}};
	\node[blue](p) at (7,8) {\voc{person}};
    \node[blue](t) at (6,5) {\voc{3rd}};
    \node[blue](sing1) at (8,5) {\voc{Singular}};
	\node[blue](a) at (9,8) {\voc{Performance}};
	\node[blue](f) at (11,8) {\voc{scope}};
    \node[blue](c) at (14,8) {\voc{question}};
    \node[blue](quant) at (13,5) {\voc{A}};
    \node[blue](polar) at (15,5) {\voc{-}};
    
	\draw[->, thick] (s.south) -- (p.north) node[midway, above, sloped] {\voc{:ARG0}};
	\draw[->, thick] (s.south) -- (c.north) node[midway, above, sloped] {\voc{:ARG1}};
 	\draw[->, thick] (s.south) -- (a.north) node[pos=0.7, sloped, above] {\voc{:aspect}};
 	\draw[->, thick] (s.south) -- (f.north) node[pos=0.65, sloped, above] {\voc{:pred-of}};
    \draw[->, thick] (c.south) -- (quant.north) node[midway, above, sloped] {\voc{:quant}};
    \draw[->, thick] (p.south) -- (sing1.north) node[midway, above, sloped] {\voc{:refer-number}};
    \draw[->, thick] (p.south) -- (t.north) node[midway, above, sloped] {\voc{:refer-person}};
    \draw[->, thick, bend left=45] 
    (f.south) to 
    node[midway, above, sloped] {\voc{:ARG0}} 
    (p.south east);
    \draw[->, thick] (f.east) -- (c.west) node[midway, above, sloped] {\voc{:ARG1}};
    \draw[->, thick] (c.south) -- (polar.north) node[midway, above, sloped] {\voc{:polarity}};
\end{tikzpicture}}
\smallbreak
\small
\begin{verbatim}
(a/ answer-01
   :ARG0 (p/ person
             :refer-person 3rd
             :refer-number Singular)
   :ARG1 (q/ question
              :quant A
              :polarity -)
   :pred-of (s/ scope
                :ARG0 p
                :ARG1 q)
    :aspect Performance)
\end{verbatim}
\caption{UMR graph for the sentence ``Someone didn't answer all the questions'' as a graph and in PENMAN notation \citep{kasper-1989-flexible}, adapted from the UMR Guidelines \citep{UMRGuidelines}.}
\label{fig:umr_example}
\end{figure}

AMR has been widely adopted in the NLP community, largely due to the extensive efforts made in developing effective text-to-AMR parsing and AMR-to-text generation models. This work has enabled AMR to be successfully applied in a variety of downstream tasks, including text summarization, machine translation, and information extraction \citep{wein-opitz-2024-survey}. 
While UMR shows initial promise in improving the performance of downstream applications for low-resource languages \citep{wein-2025-uniform}, developing text-to-UMR parsers is crucial to exploring the full scope of UMR benefits in multilingual and low-resource settings.
Without reliable parsers, UMR's richer representations cannot be extracted or leveraged in such applications, thereby limiting the utility of UMR. By building text-to-UMR parsers in English, where annotated data and pre-existing AMR models are available, we establish a foundation for transferring these models and approaches to low-resource languages.
Additionally, high-quality English AMR parsers are critical for multilingual or cross-lingual tasks leveraging UMR.

In this paper, we lay the groundwork for expanding the utility of UMR in downstream NLP applications.
Specifically, in this work we focus on  \textbf{sentence-level English UMR parsing}. We constrain our task due to the enormous amount of linguistic information captured in UMR graphs and the complicated nature of this task. Working with English provides a valuable starting point for the more challenging task of low-resource language UMR parsing, where annotated data and pre-existing text-to-AMR parsers (e.g., for Navajo or other indigenous languages) are largely unavailable. By establishing methods and baselines in English, we can better develop techniques that are likely to transfer to low-resource settings in future work.

We explore two approaches to creating text-to-UMR parsers: (1)~fine-tuning existing text-to-AMR parsers using UMR data, and (2)~converting Universal Dependency (UD) trees into partial UMR graphs following \cite{gamba-etal-2025-bootstrapping}, then training a T5 model \cite{raffel2023exploringlimitstransferlearning} to expand these partial graphs into complete UMRs. Our contributions include:
\begin{itemize}
\addtolength\itemsep{-3mm}
    \item[1.] A fine-grained analysis of the baseline pipeline's performance \citep{chun-xue-2024-pipeline} across the data in UMR v1.0 and the newer UMR dataset, UMR v2.0.
    \item[2.] Five English sentence-level text-to-UMR parser models fine-tuned from existing text-to-AMR architectures. 
    \item[3.] A fine-tuned model that converts partial UMRs to complete UMRs, employing the UD to UMR bootstrap approach by \citet{gamba-etal-2025-bootstrapping}.
\end{itemize}
Our best performing model is our fine-tuned text-to-AMR parsers, which we call SETUP: Sentence-level English-to-UMR Parser.\footnote{Our code and checkpoints are available at \href{https://github.com/ACNLPlab/UMR_Parsing.git}{https://github.com/ACNLPlab/UMR\_Parsing.git}}

\section{Background \& Related Work}
\label{sec:background}

UMR builds upon AMR, a semantic graph framework developed for English and subsequently extended to multiple languages \citep{wein-schneider-2024-assessing}. Since AMR was designed for English, each language variation of AMR has to account for language-specific features, whereas UMR’s annotation framework is designed to accommodate multilingual representations.

Furthermore, UMR allows for flexible annotation and consistency across languages by employing a lattice-like annotation structure \citep{van-gysel-etal-2019-cross}.
This annotation process permits annotators to apply either coarse- or fine-grained annotations, depending on the language's features.
Additionally, ``Stage 0'' is a part of the annotation process that allows annotators to establish predicate–argument structures and construct rolesets. This stage is employed for languages without existing rolesets, with particular emphasis on low- and no-resource languages \cite{vigus-etal-2020-cross}.
\citet{wein-bonn-2023-comparing} compare the multilingual capabilities of UMR to the cross-lingual adaptations of AMR and find that UMR more effectively captures language-specific phenomena across languages, handling a wider range of morphosyntactic and semantic constructions while maintaining a consistent, unified representation.

While both AMR and UMR capture meaning \citep{van-gysel-etal-2019-cross}, UMR extends AMR in several important ways: it represents tense, aspect, modality, and scope at the sentence level; incorporates document-level information such as temporal and modal dependencies; and facilitates the annotation of coreferential relations.
Recent work by \citet{bonn-etal-2023-mapping} highlights the key differences between AMR and UMR structures, serving as a resource to provide concrete mappings of the roles and concepts in AMR to those in UMR.

Universal Dependencies (UD; \citealp{de-marneffe-etal-2021-universal}) provide a cross-linguistically consistent syntactic framework that represents grammatical relations in a dependency-tree format.
Recent work has leveraged UD trees to bootstrap partial UMR graphs \citep{gamba-etal-2025-bootstrapping}, demonstrating that UD-informed approaches can reduce time spent on annotation and improve structural consistency across languages.
The design of UD prioritizes cross-lingual comparability, making it a valuable structural foundation that can be utilized in UMR parsing. 

Parsing techniques for other semantic representations have also been studied to various degrees. Apart from UMR and AMR, other prominent meaning representation frameworks include Universal Conceptual Cognitive Annotation, Semantic Dependencies, and Discourse Representation Structures. Universal Conceptual Cognitive Annotation (UCCA) \citet{abend-rappoport-2013-universal} is a compositional-tree-based representation, where the leaves are words and the nodes at other layers represent events and participants. The flexibility of UCCA stems from its multilayered structure allows for multi-sentence and multilingual annotation \citep{sadeddine-etal-2024-survey}. 
Parsing UCCA can be difficult due to reentrancy resulting in DAG structures, discontinuous structures and non-terminal nodes. TUPA \citep{hershcovich-etal-2017-transition} handles these issues with a transition-based neural parser by constructing the UCCA graph incrementally using a BiLSTM encoder.
More recent approaches to UCCA parsing include \citet{samuel-straka-2020-ufal}, a sequence-to-sequence model that jointly predicts nodes and edges to produce UCCA graphs within a cross-framework architecture. 
Semantic Dependencies (SDs) is an earlier family of meaning representation with the aim of going integrating semantic structure into syntactic dependency parsing \citep{oepen-etal-2014-semeval}. \citet{wang-etal-2021-automated} propose ACE, a neural structured framework that learns to combine multiple pretrained embeddings optimally. The tool applies the combinations in a graph-based parser to obtain strong performance on semantic dependency parsing.
Finally, Discourse Representation Structures (DRS) integrate first order logic into meaning representations \citep{Kamp+1984+1+42}. It is recursively structured in the form of nested boxes, equipped to encode events, semantic roles, discourse relations, and multiple sentences in a compositional framework. Logical operators are used on the boxes to signal scope. While earlier parsing again use transition-based methods \citep{evang-2019-transition}, while more recent approaches use character-level neural representations and exhibit performance gain in comparison to word-based encoders for DRS parsing \citet{van-noord-etal-2020-character}.
 
Given that UMR is a multilingual extension of AMR, prior work on AMR parsing serves as an inspiration for UMR parsing and as relevant context.
AMR parsing has seen significant advances, driven largely by the release of large-scale AMR corpora such as AMR v2.0 \citep{knight_amr_2.0} and AMR v3.0 \citep{knight_amr_3.0}.
Recent systems such as SPRING \citep{bevilacqua-etal-2021-one} and AMRBART \citep{bai-etal-2022-graph} treat AMR parsing and generation as dual tasks using pre-trained language models like BART and T5 \citep{raffel2023exploringlimitstransferlearning}. Models such as BiBL \citep{cheng-etal-2022-bibl} and LeakDistill \citep{vasylenko-etal-2023-incorporating} further refine this approach by jointly learning alignments or distilling structural information into encoder representations.
These high-performing AMR parsers serve as the foundation for extending semantic parsing to UMR, motivating our exploration of fine-tuning existing AMR parsers on UMR data to accelerate progress toward multilingual and document-level meaning representations.

Efforts towards UMR parsing have been limited to date.
First, \citet{buchholz-etal-2024-bootstrapping} use documentation resources such as lexical databases and interlinear glossed texts to generate UMR subgraphs for low-resource languages. Using this method, they correctly capture the predicate-argument structures in 109 out of 115 sentences in Arápaho. 
Next, given that mapping roles in AMR to those in UMR can be an arduous process as some individual roles can be mapped into various roles in UMR, \citet{post-etal-2024-accelerating} integrate animacy parsing and logic rules into a neural network to perform role conversion.
Their hybrid model, combining logic rules and a neural network, achieves a macro F-score of 63.5, and their model reliant on rules only achieves a weighted F-score of 76.1.
Finally, a new UMR parsing tool follows a pipeline approach converts from text to AMR, and subsequently to UMR \citep{chun-xue-2024-pipeline}. 
This tool combines models that convert text into AMR and uses mapping rules plus UD trees of sentences to generate sentence-level UMRs. Their approach integrates different models specialized in modality, temporal dependencies, and coreference to insert document-level information into the UMR by leveraging alignments between the sentences and the AMRs. They report an AnCast score of 64.2 and a SMATCH score of 72.2 for their best AMR to UMR parser at the sentence level. For document-level parsing, they achieve a comprehensive AnCast score of 61.5.

Additional work on UMR tools has focused on the task of UMR-to-text generation (the reverse direction on text-to-UMR parsing), and begun leveraging UMR for downstream tasks.
\citet{markle-etal-2025-generating} introduce the first approaches to generating text from multilingual UMR graphs, finding that fine-tuning existing AMR-to-text models yields the highest qualitative and quantitative results.
Furthermore, \citet{wein-2025-uniform} investigates UMR-based translation, highlighting how UMR's cross-lingual abstraction enhances meaning preservation across typologically diverse languages, thereby improving translation quality for low-resource languages. Collectively, these studies highlight an increasing interest in leveraging UMR for multilingual NLP tasks and downstream applications, thereby underscoring the need for reliable UMR parsers.

\section{Methods}
\label{sec:methods}

We build on prior work \citep{chun-xue-2024-pipeline}, using it as a baseline model and evaluating its performance on the newly released UMR v2.0 dataset \cite{11234/1-5902}, which will serve as a reference point for the following approaches. Next, we build on the recent UD-to-partial UMR converter \cite{gamba-etal-2025-bootstrapping} to automatically transform dependency trees into partial UMR graphs. These partial UMR graphs capture core semantic information from the text, which we then use to train a model that completes them into fully detailed UMR representations. Finally, we adopt a similar approach to that used in recent UMR-to-text generation work \citep{markle-etal-2025-generating}, where we fine-tune existing text-to-AMR parser models on UMR  data, adapting them to capture the specific structures and conventions of UMR graphs.

\subsection{Baseline Approach}
\label{ssec:baseline_method}
We use the existing UMR parser method which follows a text-to-AMR-to-UMR pipeline approach (as described in \Cref{sec:background}), as a baseline for English sentence-level UMR parsing \citep{chun-xue-2024-pipeline}. In this architecture, text–to-AMR parsers constitute the first stage of the pipeline, generating AMRs from raw sentences. These AMRs serve as intermediate semantic representations that are subsequently used as input to the UMR conversion stage.
Specifically, we use the following text-to-AMR parsing models: AMRBART \citep{bai-etal-2022-graph}, LeakDistill \citep{vasylenko-etal-2023-incorporating}, amrlib\footnote{ \href{https://github.com/bjascob/amrlib/tree/master}{amrlib GitHub Repository}}, SPRING \citep{bevilacqua-etal-2021-one}, and BiBL \citep{cheng-etal-2022-bibl}. 
Next, the resulting AMRs are passed to LEAMR \citep{blodgett-schneider-2021-probabilistic} to retrieve sentence-level AMR alignments, while UD graphs are generated automatically by the pipeline’s conversion code from the same sentences.  Following the baseline approach of \citet{chun-xue-2024-pipeline}, these alignments and UD structures together serve as inputs to the rule-based UMR conversion component constituting the final stage of the pipeline.

We test this baseline on the original UMR v1.0 dataset, as is evaluated in \citet{chun-xue-2024-pipeline}, and also evaluate it on the new UMR v2.0 dataset. The format of the new dataset requires some debugging from the pipeline model, in particular with regard to tokenization issues.\footnote{For instance, the parser code maintains ``<Architect>'' as a single token, which is a common term in the English Minecraft data of UMR v2.0; however, the UD graph expresses this as three different tokens, and therefore, the baseline code is expanded to deal with these tokenization issues.}

To get a more comprehensive baseline of current state-of-the-art UMR parsing technology, we integrate the neuro-symbolic approach to role conversion from AMR to UMR \citep{post-etal-2024-accelerating} into the pipeline.
In this process, AMR concepts are classified as animate or inanimate by the parser. Then, the model uses a combination of logical rules and a neural network along with this animacy decision to find a good UMR mapping from the original split role. 
For example, the \texttt{:source} AMR role may be mapped into \texttt{:source}, \texttt{:destination}, or \texttt{:goal} in UMR.
The model’s predictions for each role are then integrated into the corresponding UMRs produced by the pipeline, yielding the final UMRs that incorporate the neuro-symbolic role conversion approach.

\subsection{Fine-tuning Approaches}
\label{ssec:ft_method}
Next, we explore two fine-tuning approaches to achieve a text-to-UMR parsing model: (1) fine-tuning existing text-to-AMR parsers on UMR data, and (2) converting UDs into partial UMRs following \citet{gamba-etal-2025-bootstrapping}, then fine-tuning T5 \cite{raffel2023exploringlimitstransferlearning} to complete these partial graphs into full UMRs.

For the first approach, we directly fine-tune text-to-AMR parsers on UMR data for 10 epochs with a learning rate of 4e-5. The models we use include:
\begin{itemize}
\addtolength\itemsep{-3mm}
    \item[1.] amrlib: T5 model \cite{raffel2023exploringlimitstransferlearning}  trained on AMR v3.0 \cite{knight_amr_3.0}
    \item[2.] SPRING: BART-based model that treats AMR parsing and generation as two complementary, bidirectional tasks, trained on AMR v2.0 \citep{knight_amr_2.0} and AMR v3.0.
    \item[3.] BiBL: SPRING-based model that aligns AMR graphs with their respective sentences to share learning across parsing and generation, trained on AMR v2.0 and v3.0.
    \item[4.] LeakDistill: transformer-based model that uses structural adapters to capture graph information, building on word-to-node alignments trained on AMR v2.0 and v3.0.
    \item[5.] AMRBART: BART-based model trained on AMR v2.0 and AMR v3.0.
\end{itemize}

In this approach, we fine-tune the existing text-to-AMR models with UMR data, enabling the models to adapt to the UMR structures while retaining the semantic knowledge acquired during AMR training.

Next, we leverage existing work from \citet{gamba-etal-2025-bootstrapping}, who introduce a method for bootstrapping UMR graphs from UD trees.
First, we generate UDs for the sentences in the UMR datasets using Stanford's Stanza pipeline \citep{qi2020stanza}, reformat them into the CoNLL-U format, and then convert these UD trees into partial UMR graphs. Then, we train a T5 model on pairs of these partial graphs with their respective sentences and full gold reference graphs, with the objective of training the model to convert the partial graphs into complete UMR graphs. 

\subsection{Data}
\label{ssec:data}
The release of UMR v1.0 \citep{bonn-etal-2024-building} spans across six languages: English, Chinese, Sanapaná, Arápaho, Navajo, and Kukama, totaling 2,022 sentence-level UMRs. The English dataset includes fragmented transcripts of a spoken description of a silent film, as well as text from LORELEI news articles. Some examples of these include: 
\begin{itemize}
\addtolength\itemsep{-3mm}
    \item ``and he picks it up,'' 
    \item ``A-nd then . . it shifts''
    \item ``or... or before.''
\label{item:umr1_okay}
\end{itemize}
The Chinese data is comprised of sentences from Wikinews. Annotations in Arápaho, Navajo, and Kukama all correspond to narrative texts, whereas the genre of the Sanapaná annotations is not stated \cite{bonn-etal-2024-building}. 

The release of UMR v2.0 \cite{11234/1-5902} has expanded upon the initial six languages, adding Czech and Latin, and now contains 210,237 sentence-level UMRs.
However, most of these new UMRs are concentrated in two languages: English, with 29,912 graphs, and Czech, with 175,268. Notably, a large portion of the new English UMRs consists of annotated  interactions between two participants, one designated as the ``Builder'' and the other as the ``Architect,'' while they play the video game Minecraft, which is a sandbox game where players can build structures.
These UMRs capture both spoken dialogue and in-game actions, with a majority of the English sentences in the test set being Minecraft-related, such as:
\begin{itemize}
\addtolength\itemsep{-3mm}
    \item ``{[}Builder picks up a purple block at X:1 Y:2 Z:2]''
    \item ``<Architect> oops sorry, I meant behind :)''
    \item ``{[}Builder puts down a orange block at X:1 Y:2 Z:-2]''
    \item ``<Architect> We will build the `I' portion now with orange blocks''
\label{item:umr2_minecraft_sent_example}
\end{itemize}

The other portion of these new English UMRs resembles more fluid spoken English text. Examples include: 
\begin{itemize}
\addtolength\itemsep{-3mm}
    \item ``Likewise when services had to shut down after a day because of snow, the media launched a full blown attack and everyday people were saying how disgusting it was.'' 
    \item ``The percentage of lung cancer deaths among the workers at the West Groton, Mass., paper factory appears to be the highest for any asbestos workers studied in Western industrialized countries, he said.''
    \item ``Where are those people going to go when the private health insurer disappears?''
\label{item:umr2_news}
\end{itemize}

\subsection{Data Split Selection}
\label{ssec:init_exper}

\begin{table}
    \hspace{-1cm}
    \small
    \begin{center}
    \begin{tabular}{|c|c|c|c|}
    \hline
    \textbf{Model}&\textbf{AnCast}&\textbf{SMATCH}&\textbf{SMATCH++}\\
    \hline
    amrlib&12.560 & 41.287 & 43.655\\ 
    \hline
    SPRING&34.087 & 47.966 & 49.130\\ 
    \hline
    BiBL&31.393 & 49.930 & 51.960\\ 
    \hline
    LeakDistill&19.588 & 35.912 &39.086 \\ 
    \hline
    AMRBART& \textbf{47.876}& \textbf{69.289} & \textbf{69.190}\\ 
    \hline
    \end{tabular}
    \end{center}
    \caption{Overall performance of each AMR model, measured by F-scores, after fine-tuning on UMR v1.0 and comparing their predicted UMRs against the gold references. 
    }
    \label{tab:umr1_ft_results}
\end{table}

In this subsection, we describe our process for preparing the training, development, and test sets, with the goal of identifying the most effective data split for model training. This is particularly important given the composition of the datasets and their difference in makeup. As shown in \Cref{item:umr2_minecraft_sent_example}, the datasets vary considerably: UMR v1.0 contains short, fragmented spoken dialogue, while UMR v2.0 includes both Minecraft-related interactions and longer, more complex standard English text, which tend to be more structured. Due to variations in style and domain-specific vocabulary, model generalization and performance may be affected. Thus, we create and test different data splits to determine which configuration produces the most generalizable model.

We begin with the UMR v1.0 corpus \citep{bonn-etal-2024-building}, selecting 143 of the 209 English sentences that have corresponding UMR graphs. 
To prevent inadvertently rewarding the text-to-AMR models trained on overlapping sentences, we exclude 66 graphs that overlap with AMR sentences used to train the five selected text-to-AMR parser models. 

Next, using the UMR v2.0 corpus \citep{11234/1-5902}, which includes all data from UMR v1.0, we create three different splits (still excluding the 66 overlapping graphs) to evaluate the effect of different training subsets:
(1)~all UMR v2.0 data from all languages\footnote{Although our parsing approaches focus exclusively on English, including data from other languages may help the model learn more general graph structures and relations that transfer across languages.}, (2)~all English data from UMR v2.0, and (3)~English UMR v2.0 data excluding some of the repetitive Minecraft interactions.
In this third split, we retain only 1,000 sentences starting with ``[Builder'', as these typically produce highly repetitive action-descriptive sentences (examples shown in \Cref{item:umr2_minecraft_sent_example}). We exclude some of these sentences to encourage the model to generalize beyond repetitive action patterns.

We select our data split by first fine-tuning each existing text-to-AMR architecture on the English UMR v1.0 data to establish baseline performance and select the currently best-performing AMR-to-text model on that small amount of UMR data.
We then select the best-performing model, AMRBART (\Cref{tab:umr1_ft_results}), and evaluate it on three additional splits derived from UMR v2.0.
We find that AMRBART performs best on the full English UMR v2.0 split, suggesting that this configuration offers the most effective balance between data size and domain diversity.

Based on this finding, we adopt a split containing 22,938 training, 6,680 development, and 2,941 test sentences for our main experiments. 
Finally, we fine-tune all five existing text-to-AMR parsers, as well as our UD pipeline model, on this split; these constitute the models we evaluate in \Cref{sec:ft_results} and form the basis of our final experimental results.
 
\subsection{Metrics}
Graph similarity between AMR and UMR graphs can be measured with a variety of standard metrics, including AnCast \citep{sun-xue-2024-anchor}, SMATCH \citep{cai-knight-2013-smatch}, and SMATCH++ \citep{opitz-2023-smatch}. 
AnCast employs an anchor broadcast alignment algorithm designed to circumvent local maxima pitfalls. AnCast decomposes graph comparison into concept F1, unlabeled relation F1, labeled relation F1, and weighted relation F1, providing interpretable scores over nodes and edges via an efficient alignment procedure.
SMATCH measures overall graph similarity by computing the maximum F1 overlap of instance and relation triples under a variable alignment, yielding a single structural similarity score.
SMATCH++ is an extension of SMATCH with standardized preprocessing, improved alignment guarantees, and fine-grained subgraph scoring, enabling more controlled and diagnostically informative evaluation of semantic graph predictions.
We use all three of these evaluation metrics to ensure a robust understanding of parser performance.

\section{Results}
\label{sec:results}
In this section, we evaluate the performance of our different UMR parsing approaches. We first summarize the performance of the baseline pipeline, noting that it performs worse on UMR v2.0 than on UMR v1.0, due to substantial differences in the text genre. We further analyze the pipeline's performance across different splits within UMR v2.0, highlighting variations associated with data from different sources. We then report the high-scoring performance of the fine-tuning approach for both the AMR parser architectures trained on UMR data and for the UD-based approach. We evaluate our results using graph similarity metrics, including AnCast, SMATCH, and SMATCH++, and conduct a qualitative analysis of generated UMR graphs, which provides insight into how different approaches and models capture semantic and structural nuances.

\subsection{Baseline}
\label{sec:baseline_results}

\begin{table}
    \hspace{-1cm}
    \small
    \begin{center}
    \begin{tabular}{|c|c|c|c|}
    \hline
    \textbf{Model}&\textbf{AnCast}&\textbf{SMATCH}&\textbf{SMATCH++}\\
    \hline
    amrlib& 18.027 & 33.422 & 33.832\\ 
    \hline
    SPRING& \textbf{20.484}& 35.102 & \textbf{34.263}\\ 
    \hline
    BiBL& 17.808& 32.236 & 31.586\\ 
    \hline
    LeakDistill& 18.754& 35.004 & 32.992\\ 
    \hline
    AMRBART& 19.878& \textbf{35.592} & 34.078\\ 
    \hline
    \end{tabular}
    \end{center}
    \caption{For all English sentences from UMR v2.0, the overall performance of each model, measured by F-scores, after converting their generated AMRs to UMRs via the pipeline approach and comparing them against the gold UMRs.
    }
    \label{tab:overall_umr2_pipeline_results}
\end{table}

Based on the baseline model setup from \Cref{ssec:baseline_method} and the data split selected in \Cref{ssec:init_exper}, we run the pipeline approach as our baseline on all English sentences in the UMR v2.0 corpus. 
The highest AnCast score we obtain is 20.5 for SPRING, the highest SMATCH score is 35.6 for AMRBART and the highest SMATCH++ score is 34.3 for SPRING, as shown in \Cref{tab:overall_umr2_pipeline_results}. 
These UMR v2.0 scores are considerably lower than the 72.2 SMATCH score of the pipeline approach on UMR v1.0 as reported in \citet{chun-xue-2024-pipeline}.
We believe that such a difference in results is due to the stark difference between the data in the first and second UMR corpora.
As seen in \Cref{item:umr1_okay} UMR v1.0 corpus contains many sentences that are simply fragments of speech. In contrast, the second UMR corpus is completely dominated by the Minecraft sentences \Cref{item:umr2_minecraft_sent_example}. 
The AMR parsers are not equipped to handle tags indicating dialogue or coordinates tokens, which limits their performance in these contexts.

To explain the discrepancy between the results, we conduct a fine-grained evaluation of the split by evaluating the scores on the Minecraft and non-Minecraft sentences separately. We sort these by separating the UMRs based on whether they contain any ``Builder'' or ``Architect'' tags in the sentence. As expected, for the Minecraft data alone, the baseline had very low scores, with the highest scores being 13.3, 30.6, and 30.2 for AnCast, SMATCH, and SMATCH++, respectively, as shown in \Cref{tab:builder_f1_summary}.
On the contrary, the data which does not include any Minecraft narration performs better: achieving highest scores of 62.2, 64.8 and 61.3 in the same order, as shown in \Cref{tab:nobuilder_f1_summary}. 
These are standard English sentences that do not include any dialogue tags or coordinate descriptions, and are thus more in line with the AMR training data supplied to the text-to-AMR models.
Still, these standard English sentences differ from the first corpus. While UMR v1.0 is very conversational and fragmented, the UMR v2.0 non-Minecraft sentences are longer and more complex (see \Cref{item:umr2_news}), thus accounting for the smaller difference we see between the UMR v1.0 results and v2.0 results.

\begin{table}
    \hspace{-1cm}
    \small
    \begin{center} 
    \begin{tabular}{|c|c|c|c|}
    \hline
    \textbf{Model}&\textbf{AnCast}&\textbf{SMATCH}&\textbf{SMATCH++}\\
    \hline
    amrlib& 11.390& 28.171 & 29.620\\ 
    \hline
    SPRING& 12.568& 29.322 & 29.974\\ 
    \hline
    BiBL& 10.455& 26.401 & 26.886\\ 
    \hline
    LeakDistill& 11.850& 29.864 & 28.940\\ 
    \hline
    AMRBART& \textbf{13.329}& \textbf{30.600} & \textbf{30.164}\\ 
    \hline
    \end{tabular}
    \end{center}
    \caption{For English sentences \textit{with Builder/Architect tags} from UMR v2.0, the performance of each model, measured by F-scores, after converting their generated AMRs to UMRs via the pipeline approach and comparing them against the gold UMRs.}
    \label{tab:builder_f1_summary}
\end{table}

\begin{table}
    \hspace{-1cm}
    \small
    \begin{center} 
    \begin{tabular}{|c|c|c|c|}
    \hline
    \textbf{Model}&\textbf{AnCast}&\textbf{SMATCH}&\textbf{SMATCH++}\\
    \hline
    amrlib& 57.775 & 62.248 & 60.351\\ 
    \hline
    SPRING& 61.907 & 64.572 & \textbf{61.299}\\ 
    \hline
    BiBL& \textbf{62.210} & \textbf{64.843} & 61.185\\ 
    \hline
    LeakDistill& 57.010 & 61.675 & 58.492\\ 
    \hline
    AMRBART& 54.065 & 60.626 & 58.770\\ 
    \hline
    \end{tabular}
    \end{center}
    \caption{Performance of each model, measured by F-scores, after converting their generated AMRs to UMRs via the pipeline approach and comparing them against the gold UMRs for English sentences \textit{without Builder/Architect tags} from UMR v2.0. 
    }
    \label{tab:nobuilder_f1_summary}
\end{table}

Additionally, we use the neuro-symbolic approach by \citet{post-etal-2024-accelerating} to incorporate the split-role decisions of their model into the UMRs generated by the pipeline approach. The animacy parsing model is trained on five different folds of data, and so we make five different sets of UMRs with the roles predicted by the checkpoints from all folds. We obtain the scores from each of them and average them out to get the final score. However, the new scores lead to a negligible change; for example, BiBL achieves a 32.236 SMATCH score with the baseline pipeline and 32.250 after incorporating the neural conversion. As this addition has a minimal effect on overall performance, we exclude these results from further discussion.

\subsection{Fine-tuning}
\label{sec:ft_results}

In this section, we evaluate how well existing AMR parser architectures adapt to UMR parsing after being fine-tuned on UMR data. Specifically, we assess which AMR-to-text generation model performs best on the designated test split of all English data from UMR v2.0 when fine-tuned on English sentences and their respective UMR graphs, using graph similarity metrics as our primary evaluation criteria. The results in \Cref{tab:umr2_ft_summary} demonstrate that BiBL achieves the strongest overall performance among the evaluated models, with scores of 84.348, 88.816, and 90.975 on AnCast, SMATCH, and SMATCH++, respectively.

To complement our experiments with fine-tuned AMR architectures, we also evaluate our UD-based parsing approach. Interestingly, the UD approach performs similarly to the fine-tuned AMR-to-text models in terms of scores, often matching and sometimes even exceeding the results of the UMR fine-tuned SPRING and LeakDistill models.
This highlights its viability as a UD-driven method, particularly given that it is not explicitly designed for UMR. 
However, a weakness of this approach is T5’s tendency to misalign, omit, or insert extra parentheses, resulting in structurally invalid UMR graphs.
For instance, a common error of this approach is the omission of the final parenthesis in long UMR graphs. To address this, we develop a post-processing script to detect and correct parenthesis mismatches. It is also important to note that all the fine-tuning approaches outperform the baseline pipeline approach, which shows that both fine-tuning text-to-AMR parsing models and utilizing UD-driven methods show great promise as state-of-the-art UMR parsing approaches.


To better understand the strengths and weaknesses of these approaches, we examine two sets of generated UMR graphs from our fine-tuning and UD approaches, as well as the gold UMR for the same sentence.

\begin{small}
\begin{verbatim}
Sentence: <Architect> oops sorry, I meant behind :)
    
Gold UMR:
(s34a / and
 :op1 (s34s / sorry-01
        :mod (s34o / oops
              :mode expressive))
 :op2 (s34m / mean-01
        :ARG0 (s34i / i)
        :ARG2 (s34b / behind-02
                :ARG3 (s34c / cartesian-framework-91
                       :FR (s34r / relative-to-builder)))))

Fine-tuned BiBL-generated UMR:
(z0 / and
 :op1 (z1 / sorry-01
        :mod (z2 / oops
              :mode expressive))
 :op2 (z3 / mean-01
        :ARG0 (z4 / i)
        :ARG2 (z5 / behind-02
                :ARG3 (z6 / cartesian-framework-91
                       :FR (z7 / relative-to-builder))))
 :op3 (z8 / emoticon
        :value ":)"))

UD Approach-generated UMR:
(s2a / and 
 :op1 (s2s / sorry-01 
       :ARG1 (s2o / oops)) 
 :op2 (s2m / mean-01 
       :ARG0 (s2i / i) 
       :ARG2 (s2b / behind-02 
              :ARG2 s2i)))
    
\end{verbatim}
\end{small}

Analysis of the generated graphs show that both BiBL and the UD approach accurately capture the root, which is \texttt{and}, as well as the main predicates \texttt{sorry-01} and \texttt{mean-01}. BiBL nearly perfectly reproduces the gold graph, accurately representing the modifier \texttt{oops} with \texttt{:mode expressive} along with the relative structure (\texttt{:FR relative-to-builder}). In this example, BiBL achieves scores of 88.89, 91.89, and 91.89 for AnCast, SMATCH, and SMATCH++ respectively. The sole deviation is BiBL's addition of an emoticon node, which is included in the sentence and does not alter the core semantic interpretation of the graph. In contrast, the UD approach misaligns \texttt{oops} as the \texttt{:ARG1} of \texttt{sorry-01} rather than the modifier. Moreover, it entirely excludes \texttt{:mode expressive}, showing that this approach is susceptible to excluding the nuances of UMR. This is further confirmed by the automatic metrics, where this approach achieves scores of 57.143, 73.333, and 73.33 for AnCast, SMATCH, and SMATCH++, respectively. From this analysis, we find that BiBL produces UMRs that are more closely aligned with the gold reference, whereas the UD approach captures only the core meaning and is prone to omitting finer details, a pattern also reflected in the quantitative results of this example and the overall results.

\begin{small}
\begin{verbatim}
Sentence: Yes, even flowers have thorns.
    
Gold UMR:
(f / flower
    :mod (e / even)
    :ARG0-of (h / have-03
                :ARG1 (t / thorn)))

Fine-tuned BiBL-generated UMR:
(c / confirm-01
    :ARG1 (e / even-if
              :op1 (f / flower
                       :ARG0-of (t / thorn-01))))

UD Approach-generated UMR:
(a / and 
        :op1 (c / confirm-01) 
        :op2 (f / flower 
                    :mod (e / even) 
                    :ARG1-of (h / have-03 
                                :ARG0 (t / thorn))))
    
\end{verbatim}
\end{small}

In our second example, we observe a reversal of performance patterns. The UD approach more closely mirrors the gold UMR, accurately representing \texttt{even} as a modifier of \texttt{flower} via \texttt{:mod}, as well as capturing the \texttt{have-03} predicate with its correct \texttt{:ARG0}, \texttt{thorn}. However, this approach does predict the root to be \texttt{and} rather than \texttt{flower}. Furthermore, BiBL misidentifies the root as \texttt{confirm-01} and incorrectly frames \texttt{even} as a conjunction operator \texttt{even-if}, losing the modifier relationship entirely. It also fails to recover \texttt{have-03} and conflates \texttt{thorn} into \texttt{thorn-01}. This divergence from BiBL's otherwise strong performance highlights a key limitation: sentences drawn from non-Minecraft contexts, such as general declarative statements, are underrepresented in the fine-tuning data, leading to degraded performance.
Overall, we note that the imbalance between Minecraft dialogue and non-Minecraft data in the training corpus means that even the fine-tuned BiBL model is not equally well-equipped to handle all sentence types, with performance on non-Minecraft inputs remaining a limitation of the current approach.
Our models demonstrate promising initial performance on English, though our qualitative error analysis reveals avenues for future work.

\begin{table}
    \hspace{-1cm}
    \small
    \begin{center} 
    \begin{tabular}{|c|c|c|c|}
    \hline
    \textbf{Model}&\textbf{AnCast}&\textbf{SMATCH}&\textbf{SMATCH++}\\
    \hline
    amrlib& 79.393& 85.240 & 88.842\\ 
    \hline
    SPRING& 72.553& 78.990 & 81.710\\ 
    \hline
    BiBL& \textbf{84.348}& \textbf{88.816} & \textbf{90.975}\\ 
    \hline
    LeakDistill&78.108 & 81.656 & 81.465\\ 
    \hline
    AMRBART&81.700& 86.239 & 88.696\\ 
    \hline
    T5: UDPipeline&72.845& 80.624 & 82.853\\ 
    \hline
    \end{tabular}
    \end{center}
    \caption{Overall performance of each AMR model,
measured by F-scores, after fine-tuning on all English sentences from UMR v2.0 and comparing their predicted UMRs against the gold references}
    \label{tab:umr2_ft_summary}
\end{table}

\section{Conclusion \& Future Work}

In this work, we take steps towards building sentence-level English text-to-UMR parsers by exploring two primary strategies: (1)~fine-tuning existing text-to-AMR parsers on UMR data, and (2)~converting UD trees into partial UMR graphs and training a T5 model to complete them.
First, we find that the current state-of-the-art pipeline approach performs poorly on the UMR v2.0 splits due to the different nature of the data. However, as we evaluate the model on only the non-Minecraft portion of the split, we get more similar results to this approach when done on UMR v1.0.
Furthermore, we find that fine-tuning existing text-to-AMR parsers yields improved performance over the baseline, and it is our most successful approach overall. Among these text-to-AMR parsers, BiBL consistently achieves the highest scores across AnCast, SMATCH, and SMATCH++, while AMRBART ranks second. 
Interestingly, our UD approach, which leverages Universal Dependencies to create partial UMRs and then converts those partial UMRs to completes UMRs, remains competitive. It often surpasses SPRING and rivals LeakDistill, suggesting that UD-driven methods, in combination with sequence-to-sequence models, provide a viable strategy for UMR parsing.

These findings demonstrate that while UMR parsing is a challenging task due to the complexity of representing fine-grained semantic, temporal, modal, and coreference relations across sentences, we are able to achieve state-of-the-art results by leveraging existing AMR and UD tools. More broadly, our work lays the foundation for the development of UMR parsers as reliable parsing is essential for enabling UMR to be utilized in downstream applications such as machine translation, summarization, and information extraction, and to realize its promise as a cross-linguistic, meaning-based representation that extends beyond English. Focusing on English provides a practical starting point, as it offers high-quality annotated resources and AMR parsers that can be fine-tuned, allowing us to explore approaches to building models and establish strong baselines.
These insights and methods can then guide the expansion of UMR parsing to low-resource languages, where annotated data and pre-trained parsers are scarce. By establishing these foundations in English, we provide a pathway to extend UMR’s utility to multilingual and low-resource settings.

\section*{Limitations}

In order to establish strong foundational work on UMR parsing which leverages existing AMR technologies, we focus on parsing English, sentence-level UMR graphs.
As models for UMR improve and datasets expand, additional work on document-level UMR parsing for many languages will be made possible, but will continue to be a challenge for low-resource and morphosyntactically complex languages.
While we experiment with five AMR parsers, future work might explore other AMR parsers such as CLAP \citep{martinez-lorenzo-navigli-2024-efficient}, StructBART \citep{zhou-etal-2021-structure}, or MBSE \citep{lee-etal-2022-maximum}.
Future work might also extend to automatically parsing the document-level components of the UMR graphs.

AnCast++ \citep{sun-etal-2025-ancast},  is a recently introduced extension of AnCast designed to provide a unified evaluation of sentence-level graphs, modal and temporal dependencies, and coreference relations. We do not use AnCast++ in our evaluation as it was released contemporaneous to our work. 

\section*{Acknowledgments}

We thank anonymous reviewers and members of the Amherst College NLP lab for their feedback.
This work is supported by the Amherst College HPC, which is funded by NSF Award 2117377.

\section{Bibliographical References}\label{sec:reference}

\bibliographystyle{lrec2026-natbib}
\bibliography{lrec2026-example}

\section{Language Resource References}
\label{lr:ref}
\bibliographystylelanguageresource{lrec2026-natbib}
\bibliographylanguageresource{languageresource}

\end{document}